\documentclass[10pt,twocolumn,letterpaper]{article}

\usepackage{iccv}
\usepackage{times}
\usepackage{epsfig}
\usepackage{graphicx}
\usepackage{amsmath}
\usepackage{amssymb}
\usepackage{multirow}
\usepackage{booktabs}
\usepackage[super]{nth}

\usepackage[pagebackref=true,breaklinks=true,letterpaper=true,colorlinks,bookmarks=false]{hyperref}

\usepackage[dvipsnames]{xcolor}

\iccvfinalcopy 

\def\iccvPaperID{4523} 

\ificcvfinal\fi
\begin{document}

\title{Manipulation-skill Assessment from Videos with \\Spatial Attention Network}

\author{Zhenqiang Li, Yifei Huang, Yoichi Sato\\
The University of Tokyo\\
Tokyo, Japan\\
{\tt\small \{lzq, hyf, ysato\}@iis.u-tokyo.ac.jp}
\and
Minjie Cai\\
Hunan University\\
Changsha, China\\
{\tt\small caiminjie@hnu.edu.cn}
}

\maketitle

\begin{abstract}
Recent advances in computer vision have made it possible to automatically assess from videos the manipulation skills of humans in performing a task, which breeds many important applications in domains such as health rehabilitation and manufacturing. 
Previous methods of video-based skill assessment did not consider the attention mechanism humans use in assessing videos, limiting their performance as only a small part of video regions is informative for skill assessment. Our motivation here is to estimate attention in videos that helps to focus on critically important video regions for better skill assessment. In particular, we propose a novel RNN-based spatial attention model that considers accumulated attention state from previous frames as well as high-level knowledge about the progress of an undergoing task. We evaluate our approach on a newly collected dataset of infant grasping task and four existing datasets of hand manipulation tasks. Experiment results demonstrate that state-of-the-art performance can be achieved by considering attention in automatic skill assessment.

\end{abstract}

\section{Introduction}

Skill assessment is a type of evaluation often used to determine the skills and abilities a person has. Out of a variety of different skills, the assessment of manipulation skill is of particular interests and are widely used in various professional environments such as surgery, manufacturing and health rehabilitation. However, manual assessment of skill requires not only a large amount of labor and time but also expert supervision which is rare and not always available for ordinary applications. 

With recent advances in computer vision, automatic skill assessment from videos is believed to have many potential applications and begins to attract research interests in recent years \cite{Bertasius_2017_ICCV, doughty2017s, ilg2003estimation, malpani2014pairwise, pirsiavash2014assessing}. However, previous methods are either task-specific and lacking generalizability or unable to capture the fine details of motion in local regions of a video which are critically informative about the skill level. Inspired by the skill assessment procedure of a human expert during which the attention is always paid to the most critical part of a task (e.g., body pose of a basketball player rather than the basketball court during shooting), we believe it is important to capture the informative regions of a video in a general framework for more reliable skill assessment.

In this work, we propose a spatial attention-based method for skill assessment from videos. Spatial attention models, especially recurrent neural networks (RNNs) based models, have been extensively studied to capture critical areas from redundant backgrounds in video sequences since spatial attention in different frames is temporally correlated. They have been widely applied to many tasks such as action recognition \cite{du2017rpan,li2018videolstm,sharma2015action} and video classification \cite{long2018attention}. Hence, we choose to adopt RNN-based attention models to capture temporal transition patterns of attention in a video. However, existing RNN-based attention models fall short to be successfully applied for the task of skill level assessment for the following reason.

It has been observed that during the procedure of assessing the skill in a video, in addition to the visual information in each frame, human's transition of attention also depends on the region of previous attention (e.g., the attention tends to gradually transit from one object to another in adjacent frames) and the knowledge about the progress of a task (e.g., people tend to focus on different objects at different stages of a task). While previously attended regions tend to indicate which regions to attend subsequently, the high-level knowledge about the undergoing task affects when attention transits. Considering the case of assessing a basketball player's shooting skill, after focusing on the hands holding the ball for a while, the attention would be transferred to the ball trajectory and the target basket soon after the player shoots.

Therefore, to reliably estimate attention for skill assessment in videos, the attention model need consider simultaneously the following three types of information:  1) instantaneous visual information in each frame; 2) high-level knowledge about the undergoing task; 3) accumulated information of spatial attention in previous frames. 

In this work, we propose a novel RNN-based framework to estimate spatial attention for skill assessment. 
The proposed framework is mainly composed by two RNNs, one for modeling the transition patterns of spatial attention ($RNN_{att}$), and the other for modeling the progress of an undergoing task ($RNN_{task}$). 
The attention information accumulated by $RNN_{att}$ is utilized together with low-level visual information to estimate spatial attention for each video frame. The action information accumulated by $RNN_{task}$ is used to score the skill level in a video. 
In particular, the two RNNs interacts with each other: while spatial attention estimated by $RNN_{att}$ is used to focus on the informative regions for $RNN_{task}$ to better assess skill level, the accumulated action information represented by the hidden state of $RNN_{task}$ is incorporated into $RNN_{att}$ for better attention estimation.

To evaluate our approach, we use existing public datasets of hand manipulation tasks as well as a newly collected dataset which records visuomotor skills of infants at different ages (called as ``Infant Grasp Dataset"). To alleviate the difficulty of annotation, we annotate the videos pair-wisely and use a pairwise deep ranking technique for training our model. Using pairwise ranking is also a way for data augmentation. The Infant Grasp Dataset contains 4371 video pairs from 94 videos of object grasping task. Experimental results show that our proposed approach not only achieves state-of-the-art performance but also can learn meaningful attention for video-based skill assessment.

Main contributions of this paper are summarized as follows: 
\begin{itemize}
    \item We propose an attention-based method for assessing the skill level of manipulation actions from videos. To the best of our knowledge, this is the first work to incorporate attention mechanism in skill assessment.
    \item We propose a novel RNN-based spatial attention model which is carefully designed for skill assessment.
    \item We collect and annotate a new dataset for skill assessment, which records object-grasping task of infants at different ages.
    \item Extensive experiments are conducted on multiple public datasets, which not only shows that skill assessment performance could be greatly improved with attention mechanism but also validates the effectiveness of our proposed spatial attention model.
\end{itemize}

\section{Related work}

\subsection{Skill assessment}

Skill assessment has been extensively researched in the context of computer-assisted surgical training since the current process of surgical skill assessment sorely relies on subjective evaluation by human experts which is highly labor intensive. \cite{gao2014jhu, jin2018tool, malpani2014pairwise, sharma2014automated, zhang2011video, sharma2014video, zhang2015relative, zia2015automated, zia2016automated, zia2018video}. Jin \etal \cite{jin2018tool} proposed to automatically assess surgeon performance by tracking and analyzing tool movements in surgical videos, which hints us that the model for skill assessment should pay attention to the task-related regions in video rather than treat visual information in every region equally. For the purpose of automatic rehabilitation or sports skill training, there is another class of works focusing on the assessment of motion quality or sports performance through computer vision approaches \cite{Bertasius_2017_ICCV, cceliktutan2013graph, jug2003trajectory, parisi2016human, parmar2017learning, pirsiavash2014assessing, tao2016comparative, xiang2018s3d, xu2018learning}. Most of these methods are designed for specific tasks, and thus the generalizability is restricted.
Doughty \etal \cite{doughty2017s} took a step towards the general architecture for skill assessment by introducing a two-stream pairwise deep ranking framework. However, their method is purely bottom-up, without using the high-level information related to task or skill to guide the bottom-up feed-forward process. This may result in unsatisfactory performance since both task-related regions and irrelevant regions are equally treated when extracting deep features. In this work, we apply the spatial attention mechanism to guide the bottom-up feed-forward process so that irrelevant information could be filtered out and fine details that are critically informative for skill assessment could be utilized for feature generation.

\begin{figure*}
\centerline{\includegraphics[width=\textwidth]{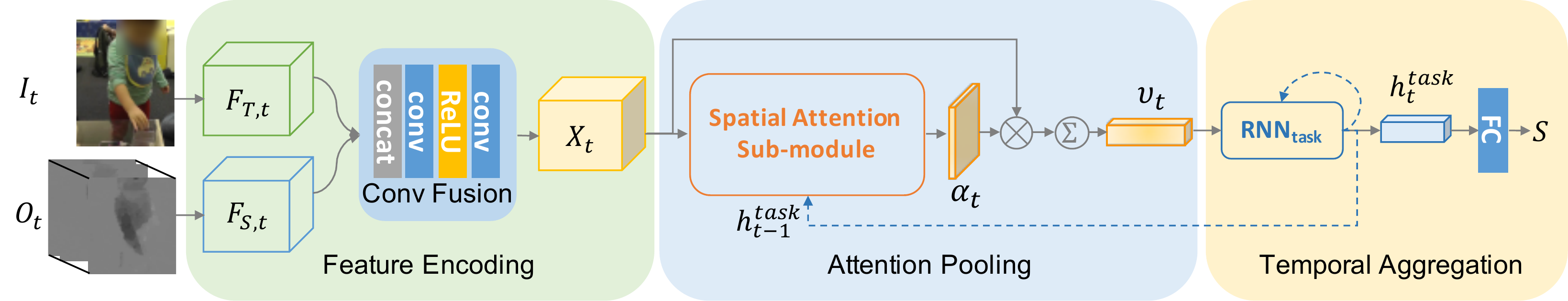}}
\caption{The illustration of our skill assessment framework. At every time step, the network takes an RGB image and the corresponding stacked optical flow images as input and firstly represents them as deep appearance-motion features. The spatial attention sub-module is then used to generate an attention map, by integrating the low-level visual information from the deep appearance-motion features and the high-level information of the undergoing task from the top part of the framework. Meanwhile, the temporal relationship between attention is also considered in this module. An attended feature vector is then generated by weighted pooling the deep feature according to the estimated attention map. The feature vector is forwarded to an RNN ($RNN_{task}$) for modeling temporal transition of actions. The output of $RNN_{task}$ at the final time step is used to yield a score for quantifying the skill level in a video.}
\label{fig:model}
\end{figure*}

\subsection{Spatial Attention mechanism}
Evidence from human perception process shows the importance of spatial attention mechanism \cite{olshausen1993neurobiological}. 
Due to its excellent performance in many image-level tasks such as image recognition \cite{wang2017residual,hu2018squeeze,park2018bam,woo2018cbam}, visual question answering \cite{das2017human, lu2016hierarchical}, image and video captioning \cite{anderson2018bottom,chen2017sca,chen2018boosted,gao2017video} and visual attention prediction \cite{huang2017temporal,huang2018predicting,huang2019mutual,selvaraju2017grad}, the spatial attention mechanism is naturally incorporated into the feature extraction process for video inputs. One of the classic applicable tasks is action recognition \cite{du2017rpan,girdhar2017attentional,li2018unified,li2018videolstm,liu2017global,ma2018attend,sharma2015action,song2017end,yan2018hierarchical}. 

The approaches of estimating spatial attention for video inputs could be divided into two classes from the perspective of whether the temporal information is involved or not. In the first class \cite{girdhar2017attentional,li2018unified,yan2018hierarchical}, the spatial attention in each frame is independently inferred without considering the temporal relationships between frames. For instance, Girdhar \textit{et al.}\cite{girdhar2017attentional} proposed an attentional pooling method based on second-order pooling for action recognition. Both saliency-based and class-specific attention are considered in their model, however, the attention is independently learned from each frame where no temporal information between frames is taken into account. It is effective in the task of action recognition which emphasizes the overall variance in appearance. However, the temporal evolution of actions performed in detailed regions is critical for skill assessment tasks so that modeling the temporal pattern becomes indispensable. The second class \cite{du2017rpan,li2018videolstm,liu2017global,ma2018attend,song2017end,sharma2015action} incorporates temporal relationships in learning spatial attention, which is usually implemented by forwarding the aggregated temporal information, i.e., the hidden states of RNNs, into the module for estimating spatial attention. Sharma \etal \cite{sharma2015action} proposed a soft-attention on top of the RNNs to pay attention to salient parts of the video frames for classification. However, when learning spatial attention for each frame, the approach only considers the high-level temporal information aggregated in previous frames without the visual information in the frame itself. Some other RNN-based methods require the auxiliary information such as human poses \cite{du2017rpan,liu2017global,song2017end} and object proposals \cite{ma2018attend} to guide the learning process of spatial attention, which not only requires the pre-computation on videos but also restricts the approach's generalization since the auxiliary information are not always available (e.g. human poses may be absent in first-person videos).

In contrast, we propose an RNN-based spatial attention module for skill assessment in this work, which fully exploits necessary information for inferring spatial attention without any pre-computed auxiliary information.


\section{Approach}
\label{sec:model}
In this section, we first describe the overview of the framework for skill assessment. Then we introduce the details of each part, especially the proposed spatial attention module which integrates temporal relationships into the estimation of spatial attention. We also describe the pairwise ranking scheme for training our model.

\subsection{Model architecture}

Our goal is to learn models for skill assessment in different tasks. Given a video in which a whole procedure of finishing a certain task is recorded, our model estimates a score to assess the skill performed in the video. Figure \ref{fig:model} depicts the architecture of our model. 
As is done in \cite{doughty2017s}, we split the video into $N$ segments, and randomly sample one frame in each segment to form a sparse sampling of the whole video. At every time step $t \in \{1, \cdots, N\}$, the \textbf{feature encoding module} extracts deep appearance-motion features from a single RGB image and the corresponding stacked optical flow images. To estimate a spatial attention map, the \textbf{attention pooling module} accepts both the deep appearance-motion features and a task-related feature generated from the temporal aggregation module which will be described below. 
The temporal relationship between attention is also considered in this module to capture the transition pattern of attention. By pooling the deep appearance-motion features with weights derived from the attention map, an attended vector is generated to encode the actions that are critically informative about the skill level. The vector is then fed into the \textbf{temporal aggregation module}, which aggregates the action information carried by the vector temporally and tries to model the temporal transition of actions in a certain task. A score for quantifying skill level in a video is regressed from the final accumulated action information. We illustrate the details of each module in the following subsections.

\begin{figure*}
    \centerline{\includegraphics[width=\textwidth]{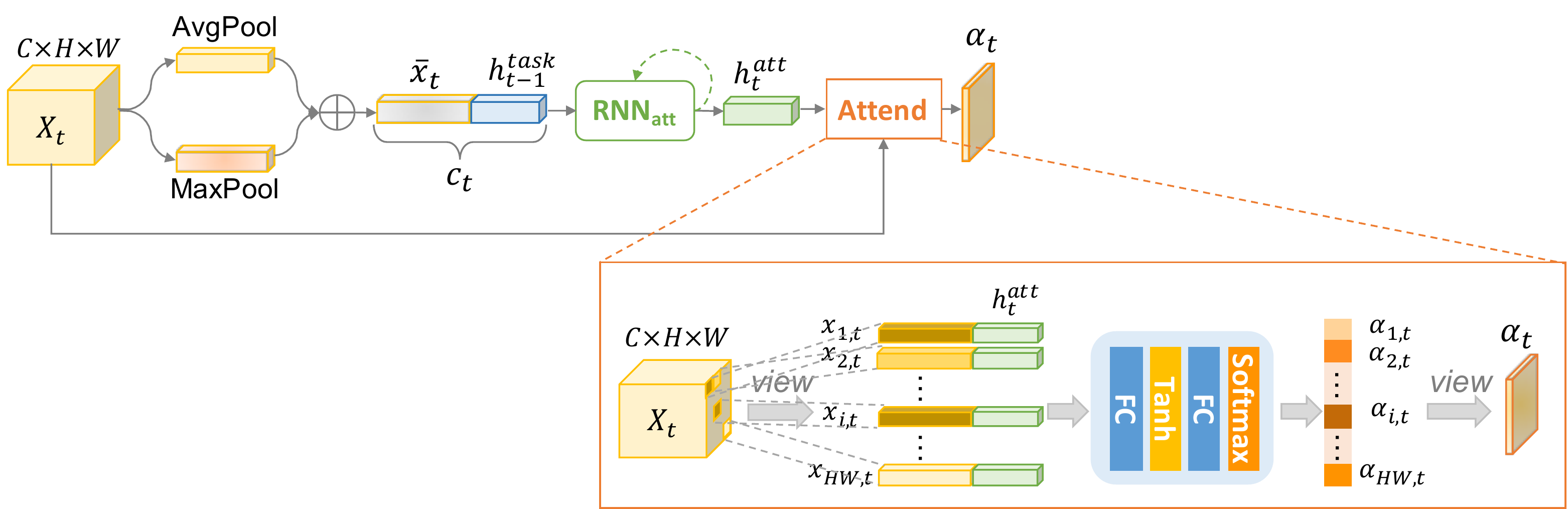}}
    \caption{The details of the spatial attention sub-module. To estimate the task-related significance for the regional vectors at different locations, the module incorporates not only the low-level visual information $\bar{x}_t$ globally extracted from the deep feature maps, but also the high-level information of the undergoing task $h_{t-1}^{task}$ which is accumulated by a high-level RNN ($RNN_{task}$ in Figure \ref{fig:model}). We also use an RNN ($RNN_{att}$) to accumulate the attention information and learn the temporal relationship of attention. The hidden state of $RNN_{att}$ is utilized to estimate attention weights for all locations in deep feature map $X_t$.}
    \label{fig:spatial}
\end{figure*}

\subsection{Feature encoding}
From each of the $t$-th segment, the module takes an RGB image $I_t$ and the corresponding stacked optical flow images $O_{t}$ as input, since appearance and motion are both important for skill assessment. The feature encoding module first extracts two deep features $F_{S,t}$ and $F_{T,t}$ from $I_t$ and $O_t$ by feeding them into two ResNet101 networks respectively. The ResNet101 networks are pre-trained on ImageNet \cite{deng2009imagenet} and then fine-tuned on UCF101 \cite{soomro2012ucf101} in a two-stream framework for action recognition\cite{simonyan2014two}. As in \cite{girdhar2017actionvlad, huang2018predicting}, we extract deep features from the last convolution layer of the 5th convolutional block. 

Then we build a two-layer convolution network for feature fusion, which takes $F_{S,t}$ and $F_{T,t}$ as input and outputs the fused deep appearance-motion representation $X_{t}\in \mathbb{R}^{C\times H\times W}$.
\begin{equation}
X_t = f_{conv2}(ReLU(f_{conv1}([F_{S,t};F_{T,t}]))).
\end{equation}

\subsection{Attention pooling}
We aim to apply attention to guide the bottom-up feed-forward process. In our work, this is done by our proposed attention pooling module that dynamically adjusts spatial attention based on both the low-level visual information and the high-level knowledge about the undergoing task. 
Specifically, at each time step, the attention pooling module accepts two inputs: deep appearance-motion feature maps of the frame as the low-level visual information, and the hidden state vector given by top RNN in temporal aggregation module representing the high-level knowledge about the undergoing task. The two inputs will be explained in detail as follows. 

To extract a compact low-level representation vector from the deep appearance-motion feature maps, following \cite{woo2018cbam}, we firstly squeeze spatial information of the deep feature map $X_t$ by performing average-pooling and max-pooling. As a result, two features are generated and summed together to form a highly abstract low-level representation vector $\bar{x}_t\in \mathbb{R}^{C}$ for $X_t$ as following:
\begin{equation}
\bar{x}_t=AvgPool(X_t) + MaxPool(X_t).
\end{equation} 
The high-level representation vector $h_{t-1}^{task}$ and the low-level representation vector $\bar{x}_t$ both serve as a basis for estimating attention map. We name this part as \textit{spatial attention sub-module}, and its details are shown in Figure \ref{fig:spatial}. Briefly speaking, we concatenate the two vectors together, deriving a vector $c_t$ integrating the information for estimating the attention map:
\begin{equation}
c_t = Concat[\bar{x}_t; h^{task}_{t-1}].
\end{equation} 

Since the attention in video is temporally correlated, a Recurrent Neural Network (RNN) \cite{chung2014empirical} (called $RNN_{att}$) is adopted to capture the transition pattern of attention by temporally accumulating the attention information $c_t$. The output $h_{t}^{att}$ at time step $t$ is a vector integrating both the current low-level visual information and the high-level knowledge about the undergoing task. Moreover, the temporal dependencies between the information for attention estimation are also taken into account simultaneously:
\begin{equation}
h^{att}_t = RNN_{att}(c_t).
\end{equation}

Given the output vector $h^{att}_t$ and the deep feature maps $X_t$, our model generates an attention weight $a_{i,t}$ for each spatial location $i$ of the deep features $x_{i,t}$ at all $H\times W$ locations of feature maps and normalize them by $softmax$ activation function. The output of $softmax$ activation function is marked as $\alpha_{i,t}$. With this procedure, the attention on each spatial location will be guided by the integrated information $h_t^{att}$, which leads to a better decision on the importance of each specific location.
\begin{equation}
\begin{split}
a_{i,t}=\omega^T_a[tanh(W_{xa}x_{i,t}&+b_{xa}+W_{ha}h^{att}_t+b_{ha})], \\
& i=1,2,...,H \times W.
\end{split}
\end{equation}
\begin{equation}
\alpha_{t} = softmax(a_{t}).
\end{equation}
We call this part as \textit{Attend} part in the spatial attention sub-module (Figure \ref{fig:spatial}).

The attended image feature which will be used as input to the final RNN for skill determination is calculated as a convex combination of feature vectors at all locations:
\begin{equation}
v_{t}=\sum_{i=1}^{H\times W}\alpha_{i,t}x_{i,t}.
\end{equation}

\subsection{Temporal aggregation}
In \cite{doughty2017s,simonyan2014two,wang2016temporal}, a video-level prediction is derived by averaging the image-level predictions of sampled frames, in which no temporal relationship between information in different frames is considered. However, in skill assessment, it is hard to yield an accurate prediction only from the visual information in a single frame, since the skill level is determined by the temporal evolution of actions. For example, during the procedure of assessing one task, both the execution order of different actions and the speed of performing an action could affect the judgment of the skill level. For this reason, we choose to use an RNN to model the temporal transition of actions in a certain task by accumulating the changed action information temporally. The score for skill level will be estimated based on the output of this RNN at the final step. 

Specifically, we aggregate the feature vectors temporally using an RNN (noted as $RNN_{task}$),
\begin{equation}
h^{task}_t=RNN_{task}(\left\{v_1,v_2,\cdot\cdot\cdot,v_t\right\}). \\
\end{equation}
The output at the final temporal step of $RNN_{task}$ is forwarded into a fully connected layer (FC) to get the final score $S \in \mathbb{R}$ for quantifying the skill level in a video:
\begin{equation}
S=FC(h^{task}_N).
\end{equation}
Here the hidden state vector $h_t^{task}$ is seen as representation of the knowledge about the undergoing task since it accumulated the action information in previous frames.

\begin{figure*}
\centerline{\includegraphics[width=0.85\textwidth]{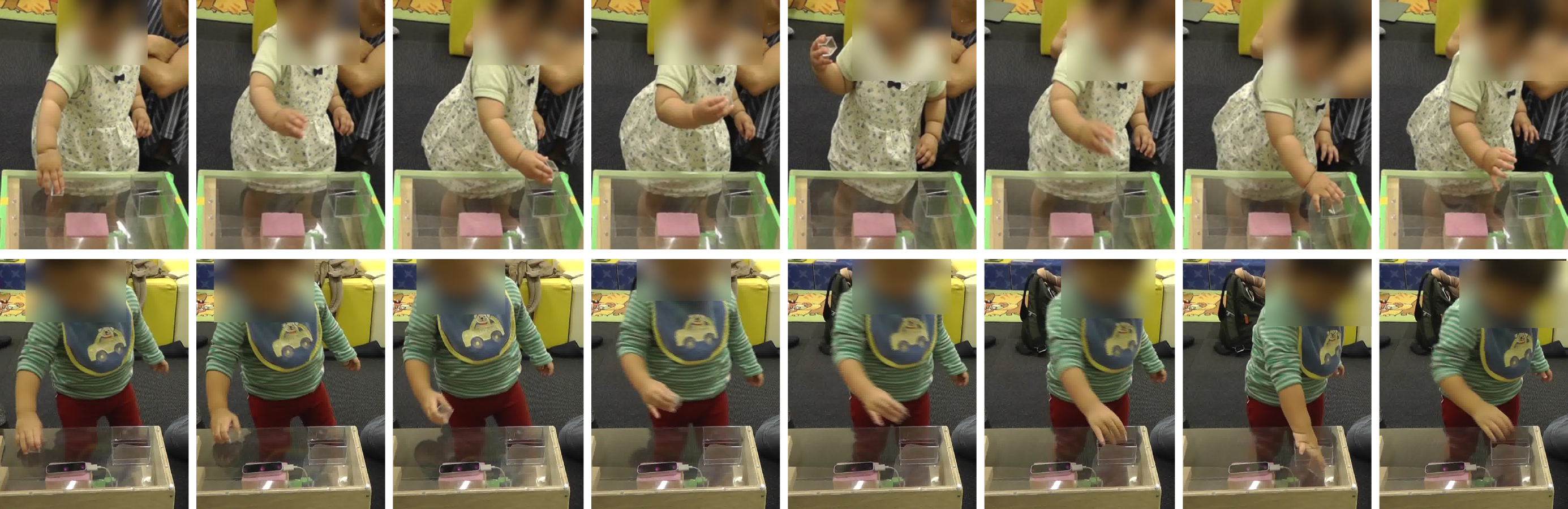}}
\caption{Image examples of two videos showing different skill levels in our Infant Grasp Dataset. The skill level in the bottom row is better than in the top row because the action of putting is not continuous in the top row. Infants' faces are blurred for privacy.}
\label{fig:Dataset}
\end{figure*}

\subsection{Training and implementation details}
We use a pairwise ranking framework \cite{doughty2017s, yao2016highlight} for training, which requires only pairwise annotations to assess the skill in videos of the same actions. To be more precise, given $M=m(m-1)/2$ pairs of videos formed from a set of $m$ videos $\{V_1, V_2, \cdots,  V_m\}$, annotators only need to point which video shows better skill in each pair rather than giving an exact score for each video:
\begin{equation}
P(V_i, V_j) = 
\begin{cases}
1 &V_i \text{ performs better than } V_j,\\
-1 &V_j \text{ performs better than } V_i,\\
0 &\text{no skill disparity.}
\end{cases}
\label{eq:Expertise}
\end{equation} 
Since $P(V_i, V_j) = - P(V_j, V_i)$, we can change the order of videos in each pair to ensure $P{\geq}0$. Only video pairs performing skill disparity are chosen for training and testing. Therefore, denote $\Psi$ as all pairs of videos in the training set that contains skill disparity, all the video pairs $\left<V_i, V_j\right> $ in $\Psi$  will satisfy  $\forall V_i, V_j \in \Psi, P(V_i, V_j)=1$. In our pairwise ranking framework, two videos in one pair are fed into a Siamese architecture consisting of two same models with shared weights.
The output of each model is a score denoted as $S(\cdot)$, and the model learns to minimize the following loss function:
\begin{equation}
L = \sum_{V_i,V_j\in \Psi}max(0, -S(V_i)+ S(V_j)+ \epsilon).
\end{equation}
$S(V_i)$, $S(V_j)$ depict the predicted skill measure for videos $i$ and $j$ respectively. $\epsilon$  denotes margin, which is incorporated to adjust the distance between the predicted scores of the two videos. In this work, we empirically select $\epsilon=0.5$ in all experiments. What should be noticed is that the score $S(\cdot)$ is not an absolute score given by annotators but learned from the training data such that the scores are consistent with pair-wise rankings given as ground truth. The loss function will punish the model if the predicted score $S(V_i)$ is no larger than $S(V_j)$ by $\epsilon$ for the video pair $\left<V_i, V_j\right> $ in which $V_i$ has higher skill level.

We use PyTorch \cite{paszke2017automatic} to implement our framework. The optical flow images for motion input are extracted by TV-$L^{1}$ algorithm \cite{zach2007duality}. For the dataset of Infant-Grasp, the optical flow images are extracted with the original frame rate and for the other datasets, we use the frame rate of 10-fps since motion is slow in these videos. The deep features in Figure \ref{fig:model} is extracted from the output of the 5-th convolution block ($conv5\_ 3$) of ResNet101 \cite{he2016deep}. The input images are resized to $448\times448$, so the size of deep features extracted from ResNet is $2048\times14\times14$. The conv-fusion module consists of 2 convolution layers, in which the first layer followed by ReLU activation. The first layer has 512 kernels with a size of $2\times2$, and the second convolution layer has a kernel size of 1 with 256 output channels. The dimensions of parameters $\left\{\omega_{a}, W_{xa}, b_{xa}, W_{ha}, b_{ha}\right\}$ in the Attend part of the spatial attention sub-module are set as $\left\{1\times32, 32\times512, 32\times1, 32\times128, 32\times1\right\}$. The RNN for both attention pattern learning and temporal aggregation are implemented with a 1-layer Gated Recurrent Unit (GRU) \cite{chung2014empirical} whose hidden state size is set as 128. We uniformly split each video into $N=25$ segments, and sample one frame randomly from each segment during training. The last frame of each segment is utilized to test our model. We use stochastic gradient descent with a momentum of 0.9 to optimize our model. We set learning rate as 5e-4 for the Infants-grasp dataset, and 1e-3 when for other datasets. All weight decays are set as 1e-3.
As \cite{doughty2017s}, our model is trained and tested separately on different datasets.
\section{Experiments}
We evaluate our method on our newly collected dataset as well as four public datasets. Similarly to \cite{doughty2017s}, we report the results yielded by four-fold cross-validation, and for each fold, we use ranking accuracy as the evaluation metric. Ranking accuracy is defined as the percentage of correctly ranked pairs among all pairs in the validation set.
\subsection{Datasets}

\begin{figure*}
\centerline{\includegraphics[width=0.85\textwidth]{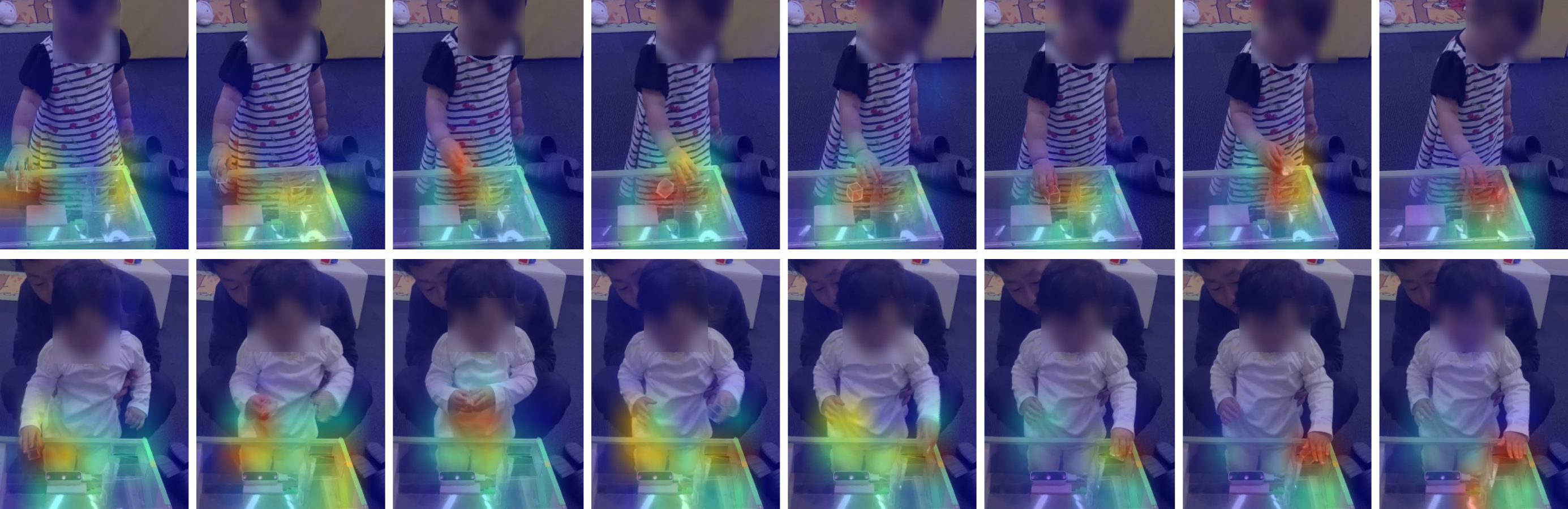}}
\caption{Visualization of attention maps estimated by our method on the Infant Grasp dataset.}
\label{fig:infant_grasp}
\end{figure*}

\subsubsection{Infant Grasp Dataset}
Since the related public datasets are either small in size (e.g., up to 40 videos \cite{doughty2017s, gao2014jhu}) or unsuitable for manipulation skill assessment (e.g., comparing skill between different diving actions \cite{pirsiavash2014assessing}), we construct a larger dataset for infant grasp skill assessment. The dataset consists of 94 videos, and each video contains a whole procedure of an infant grasping a transparent block and putting it into a specified hole. The videos were originally captured for analyzing visuomotor skill development of infants at different ages. Figure \ref{fig:Dataset} shows representative frames selected from a pair of videos. The length of each video ranges from 80 to 530 frames with a frame rate of 60fps. This dataset is expected to be of great importance not only to the computer vision community but also to the developmental psychology community.
To annotate the dataset, we asked 5 annotators from the field of developmental psychology to label each video pair by deciding which video in a pair shows a better skill than the other or there is no obvious difference in skill. We form 4371 pairs out of 94 videos, among which 3318 pairs have skill disparity (76\%).

\subsubsection{Public datasets}
We also evaluate our method using public datasets of another four manipulation tasks: Chopstick-Using, Dough-Rolling, Drawing \cite{doughty2017s}, and Surgery \cite{gao2014jhu}. The Chopstick-Using dataset contains 40 videos with 780 total pairs. The number of pairs of video with skill disparity is 538 (69\% of total pairs). The Dough-Rolling dataset selects 33 segments about the task of pizza dough rolling from the kitchen-based CMU-MMAC dataset\cite{de2008guide} and 538 pairs of videos are annotated with skill disparity (69\%). The Drawing dataset consists of two sub-dataset and 40 videos in total, among which 380 pairs are formed and 247 pairs show skill disparity (65\%). The Surgery dataset contains three sub-datasets of three different kinds of surgery tasks: 36 videos of Knot-Tying task, 28 videos of Needle-Passing and 39 videos of Suturing. Each sub-dataset contains a maximum of 630, 378, 701 pairs respectively, and since the annotation is given by a surgery expert using a standard and structured method, more than 90\% of pairs contains the difference in skill level. Following \cite{doughty2017s}, we train and test the 3 sub-datasets of the Surgery dataset together using one model. Same is done for the Drawing dataset.

\subsection{Baseline methods}
We compare with several baseline methods to validate the effectiveness of our proposed approach.

We firstly compare our method with \textbf{Doughty \textit{et al.}} \cite{doughty2017s} which is the most relevant work with ours. They use the TSN \cite{wang2016temporal} with a modified ranking loss function for skill assessment.

We also use four more baseline methods by replacing the spatial attention module in our method with existing spatial attention modules so as to evaluate the effectiveness of our spatial attention module.
\begin{itemize}
    \item \textbf{Attention Pool\cite{girdhar2017attentional}}: A static attention model originally proposed for action recognition task. 
    The spatial attention is estimated independently for each frame.
    \item \textbf{CBAM Attention\cite{woo2018cbam}}: Another static attention model which is proposed for any CNN-based tasks. Both channel and spatial attention are estimated. Similar to \cite{girdhar2017attentional}, the attention is estimated independently for each frame with low-level visual information.
    \item \textbf{Visual Attention\cite{sharma2015action}}: An RNN-based spatial attention model originally proposed for action recognition task. Only the state information of RNN for action recognition is explicitly used to estimate spatial attention. 
    \item \textbf{SCA-CNN \cite{chen2017sca}}: An RNN-based attention model originally proposed for image captioning task. Both the visual information in each frame and the state information of RNN for image captioning are exploited. 
\end{itemize}
The implementation details of these baseline attention models are further described in the supplementary. Except for the spatial attention module, all the other parts of our framework remain invariant when we are composing baseline attention methods. 


\begin{figure*}
\centerline{\includegraphics[width=0.8\textwidth]{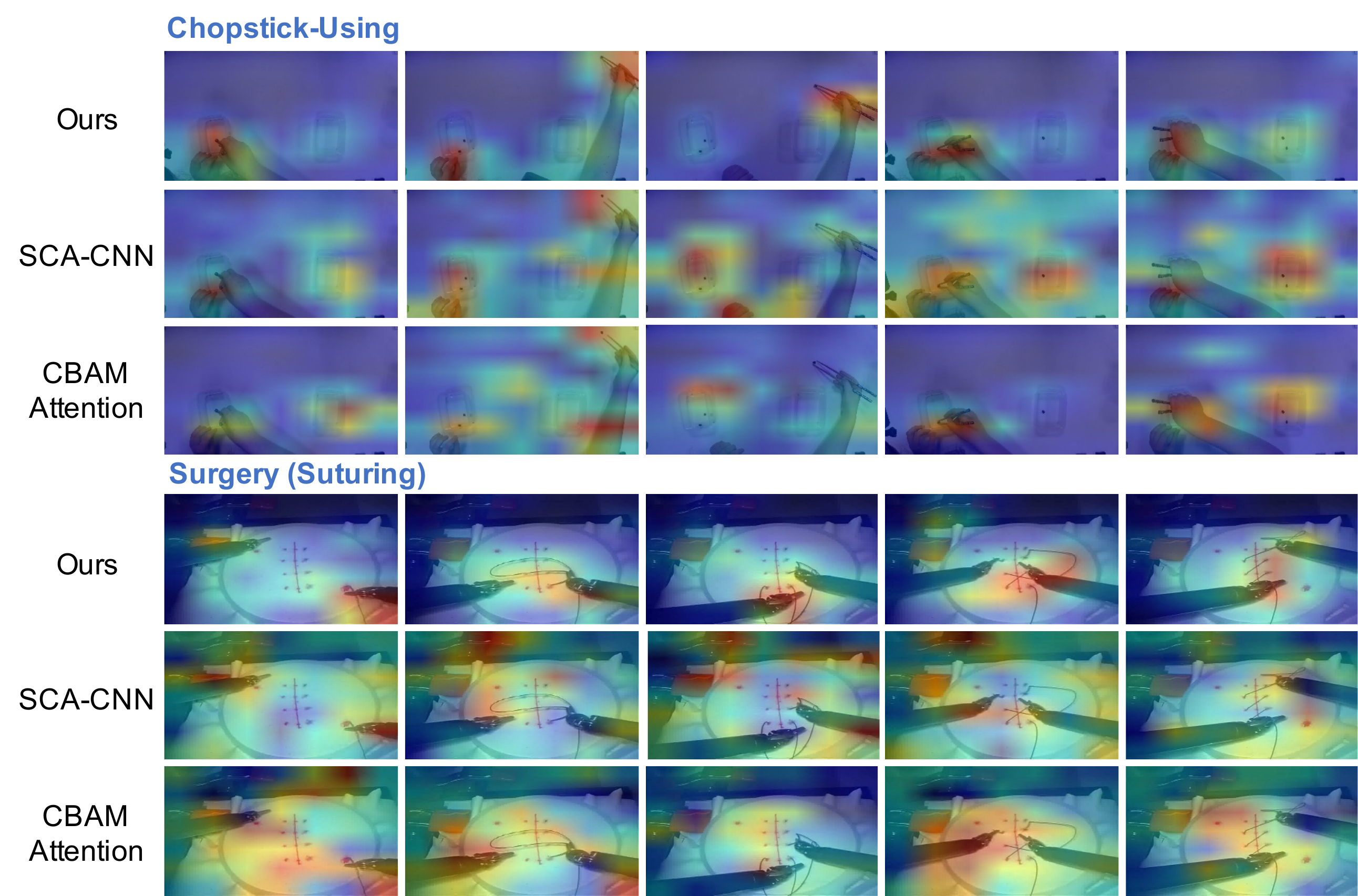}}
\caption{Visualization of the estimated attention maps estimated by our method, SCA-CNN \cite{chen2017sca} and CBAM Attention \cite{woo2018cbam} on the Chopsticks-Using and Surgery (Suturing) dataset. It can be seen that comparing with the two baselines, our method successfully focuses on the critical regions that is informative for skill assessment.}
\label{fig:result1}
\end{figure*}

\subsection{Performance comparison}

Quantitative results of different methods are shown in Table \ref{table:baseline}. Our method achieves the best performance on all datasets and outperforms the state-of-the-art method \cite{doughty2017s} by a large margin. 
Comparing within four baseline attention models, overall RNN-based attention models \cite{sharma2015action,chen2017sca} outperform static attention models \cite{girdhar2017attentional,woo2018cbam}. Our method outperforms all baseline attention models, which validates the effectiveness of our proposed spatial attention module for skill assessment.

\begin{table}
\centering
\scalebox{0.75}{
\begin{tabular}{lccccc}
\toprule
\multicolumn{1}{c}{Accuracy(\%)} &\begin{tabular}[c]{@{}c@{}}Chopstick-\\ Using\end{tabular} & Surgery & Drawing & \begin{tabular}[c]{@{}c@{}}Rough-\\ Rolling\end{tabular} & \begin{tabular}[c]{@{}c@{}}Infant-\\ Grasp\end{tabular}  \\ 
\midrule
Doughty \textit{et al.} \cite{doughty2017s}& 71.5    & 70.2    &83.2     &79.4      &80.3        \\
\midrule
Attention Pool \cite{girdhar2017attentional} &74.9  &68.8  &83.9  &79.3    &83.6   \\
CBAM Attention \cite{woo2018cbam} &82.0   &68.6    &84.1     &78.9     &83.8    \\
Visual Attention \cite{sharma2015action} &84.6   &68.1   &83.1    &79.8    &84.9   \\
SCA-CNN \cite{chen2017sca} &84.2   &69.6    &84.4     &81.8     &84.6  \\
Ours  &\textbf{85.5}   & \textbf{73.1}    &\textbf{85.3}   &\textbf{82.7}      &\textbf{86.1}   \\ 
\bottomrule
\end{tabular}
 }
\vspace{0.1cm}
\caption{Performance comparison with baseline methods. Ranking accuracy is used as the evaluation metric.}
\label{table:baseline}
\end{table}


We visualize the attention maps generated by our model. Figure \ref{fig:infant_grasp} shows attention maps on our Infant-Grasp dataset. To illustrate the transition process of the generated attention, we select eight frames for each video, in which nearly all the key actions could be captured. The performance shown in the two videos is in relatively low skill level since the cube is dropped or switched between hands. However, it can be seen that the generated spatial attention focuses on the detailed regions in all images which are critically informative for skill assessment. In the first row, the spatial attention smoothly shifts between the infant's hand and the dropped cube, instead of locating on hand regions continuously. In the second row, our attention module successfully locates the correct task-related hand when the cube is switched between two hands. The reason why our attention model is adaptive to the shifting attention might be owing to the incorporation of knowledge about the undergoing task and the modeling of temporal relationship between attention.

We also compare the attention maps of two different datasets obtained by our proposed model and two baseline attention models of SCA-CNN \cite{chen2017sca} and CBAM Attention \cite{woo2018cbam} in Figure \ref{fig:result1}. With our model, the attention is always paid onto the important regions that is informative for skill assessment. 
In contrast, the models of SCA-CNN and CBAM Attention tend to locate the salient but task-unrelated regions. For CBAM Attention \cite{woo2018cbam}, the reason might be the absence of temporal information. For SCA-CNN \cite{chen2017sca}, the reason might be the lack of explicit consideration of temporal relationship between attention.

\subsection{Ablation study}
\label{ssection_ablation}

To validate the effectiveness of each component of our model, we conduct ablation study on each dataset with the following baselines:
\begin{itemize}
    \item No Attention: Spatial attention module is entirely removed and visual information $AvgPool(X_t)$ is directly forwarded into $RNN_{task}$. We build this baseline to examine the effectiveness of spatial attention in skill assessment.
    \item No $RNN_{att}$: In spatial attention module, $RNN_{att}$ is replaced by one fully-connected layer. We build this baseline to examine the effectiveness of attention transition patterns learned by $RNN_{att}$.
    \item $\bar{x}_t$ based Attention: The $RNN_{att}$ takes only $\bar{x}_t$ as input without the concatenation with $h^{task}$. We build this baseline to examine the effectiveness of low-level visual information.
    \item $h^{task}$ based Attention: The $RNN_{att}$ takes only $h^{task}$ as input. We build this baseline to examine the effectiveness of the high-level knowledge about undergoing task.
\end{itemize}
Note that except for $No Attention$, all other baselines adopt attention mechanism for skill assessment.

\begin{table}
\centering
\scalebox{0.75}{
\begin{tabular}{lcccccc}
\toprule
\multicolumn{1}{c}{Acc(\%)} &\begin{tabular}[c]{@{}c@{}}Chopstick-\\ Using\end{tabular} & Surgery & Drawing &\begin{tabular}[c]{@{}c@{}}Rough-\\ Rolling\end{tabular} & \begin{tabular}[c]{@{}c@{}}Infant-\\ Grasp\end{tabular}\\
\midrule
No Attention             &82.1    &68.3    &82.8   &77.3        &84.0\\
No $RNN_{att}$        &84.1    &70.8    &82.4     &82.0    &85.1\\
$\bar{x}_t$-based Attention &84.1    &70.1    &83.4     &81.8    &85.3\\
$h^{task}$-based Attention &82.8    &69.1    &84.8     &81.6    &84.7\\ 
Full model            &\textbf{85.5}   &\textbf{73.1}  &\textbf{85.3}   &\textbf{82.7}   &\textbf{86.1}\\ 
\bottomrule
\end{tabular}
}
\vspace{0.1cm}
\caption{Ablation study for different components of our model. Ranking accuracy is used as the evaluation metric.}
\label{table:ablation}
\end{table}

Ablation study results are shown in Table \ref{table:ablation}. The performance of $No Attention$ is worse than other attention-based baselines, indicating the necessity of adopting attention mechanism for skill assessment. The advantage of our full model over other attention-based baselines validates our thought that the three types of information about instantaneous visual information ($\bar{x}_t$), high-level knowledge of the undergoing task ($h^{task}$), and the accumulated information of attention in previous frames ($RNN_{att}$) are all important elements for attention-based skill assessment.

\section{Conclusion}
\label{sec:conclusion}

In this study, we firstly incorporate the spatial attention mechanism into the task of skill assessment. Rather than merely duplicating of existing attention estimation models used in other computer vision tasks, we specially designed our spatial attention module for skill assessment by considering three elements: 1) instantaneous visual information in each frame (deep appearance-motion features); 2) high-level task-related knowledge; 3) accumulate information of attention. A new dataset was collected and annotated, which contains a larger number of videos recording infants' grasping skills. Experiments on multiple public datasets demonstrate that our proposed method achieves state-of-the-art performance on all datasets.


{\small
\bibliographystyle{ieee}
\bibliography{egbib}
}

\end{document}